%% file: main-arxiv.tex
\documentclass[final]{cvpr}

\usepackage{times}
\usepackage{epsfig}
\usepackage{graphicx}
\usepackage{amsmath}
\usepackage{amssymb}
\usepackage[table]{xcolor}%
\usepackage{adjustbox}

\usepackage{algorithm, algorithmic}
\usepackage[toc,page]{appendix}
\usepackage{arydshln} %
\usepackage{bm}
\usepackage{booktabs} %
\usepackage{caption}
\usepackage{cellspace}
\usepackage{color, colortbl}  %
\usepackage{comment}
\usepackage{cuted}
\usepackage{hhline}
\usepackage{makecell}
\usepackage{multirow}  %
\usepackage{subcaption}
\usepackage{soul}
\usepackage{xcolor}

\DeclareCaptionFormat{myformat}{#3}
\definecolor{ffe1da}{RGB}{255,225,218}
\definecolor{F7E0D5}{RGB}{247,224,213}
\definecolor{darkF7E0D5}{RGB}{209,154,128}
\definecolor{noel_color}{RGB}{10, 114, 72}

\usepackage{pifont}%
\newcommand{\xmark}{\ding{55}}%

\newcommand{\methodName}{\textsc{SelfMask}\xspace}

\newcommand{\tit}[1]{\textit{#1}}
\newcommand{\p}[1]{\left(#1\right)}

\newcommand{\mcal}[1]{\mathcal{#1}}
\newcommand{\mbb}[1]{\mathbb{#1}}
\newcommand{\mbf}[1]{\mathbf{#1}}
\newcommand{\mrm}[1]{\mathrm{#1}}

\newcommand{\kmeans}{$k$-means\xspace}

\newcommand{\framingPrior}{framing\xspace}
\newcommand{\distinctivenessPrior}{distinctiveness\xspace}

\usepackage[pagebackref=false,breaklinks=true,colorlinks,bookmarks=false]{hyperref} %
\usepackage{cleveref}

\def\confYear{CVPR 2022}

\begin{document}

\title{Unsupervised Salient Object Detection with Spectral Cluster Voting}

\author{Gyungin Shin$^{1}$ \quad \quad \quad Samuel Albanie$^2$ \quad \quad \quad Weidi Xie$^{1,3}$\\ [2pt]
$^1$ Visual Geometry Group,  University of Oxford, UK\\
$^2$ Department of Engineering,  University of Cambridge, UK\\
$^3$ Shanghai Jiao Tong University, China\\
{\tt\small gyungin@robots.ox.ac.uk}\\
}

\maketitle

\input{sections/00-abstract}
\input{sections/01-introduction}

\input{sections/02-related}

\input{sections/03-method}

\input{sections/04-experiments}
\input{sections/05-conclusion}

\input{sections/07-acknowledgement}

{\small
\bibliographystyle{ieee_fullname}
\bibliography{refs}
}
\clearpage
\input{sections/06-supp-mat}

\end{document}

%% file: sections/00-abstract.tex
\begin{abstract}
In this paper, we tackle the challenging task of unsupervised salient object detection (SOD) by leveraging spectral clustering on self-supervised features. 
We make the following contributions:
(i)~We revisit spectral clustering and demonstrate its potential to group the pixels of salient objects;
(ii)~Given mask proposals from multiple applications of spectral clustering on image features computed from various self-supervised models, e.g., MoCov2, SwAV, DINO,
we propose a simple but effective winner-takes-all voting mechanism for selecting the salient masks, leveraging object priors based on \framingPrior and \distinctivenessPrior;
(iii)~Using the selected object segmentation as pseudo groundtruth masks, 
we train a salient object detector, dubbed \methodName, 
which outperforms prior approaches on three unsupervised SOD benchmarks.
Code is publicly available at
\href{https://github.com/NoelShin/selfmask}{\tt https://github.com/NoelShin/selfmask}
\end{abstract}

%% file: sections/01-introduction.tex
\section{Introduction}

\textit{Salient object detection}~(SOD)\footnote{
In contrast to \textit{object detection} (which aims to localise and recognise objects with bounding boxes), \textit{salient object detection} aims to segment foreground objects by predicting pixel-wise masks for them.}, 
which aims to group pixels that attract human visual attention,
has been extensively studied in the field of computer vision due to its wide range of applications such as photo cropping~\cite{suh2003uist, marchesotti2009iccv}, re-targeting in images~\cite{ding2011cvpr, sun2011iccv} and video~\cite{Rubinstein2008tog}. 

In the literature, 
early work tackled this problem by utilising low-level features (e.g., colour~\cite{ming2015tpami}) together with priors on salient regions in an object such as contrast priors~\cite{itti1998tpami}, 
boundary priors~\cite{wei2012eccv} and centre priors~\cite{judd2009iccv}.
Recent SOD models have approached this task from the perspective of representation learning, typically training deep neural networks (DNNs) on a large-scale dataset with manual annotations.
However, the scalability of such supervised learning approach is limited because it is a costly process to collect ground-truth mask annotations. 

To overcome the necessity of large-scale human annotation, 
many unsupervised methods for saliency detection/object segmentation have recently been proposed~\cite{bielski2019emergence, chen2019unsupervised, voynov2020unsupervised, voynov2021obj, melas2021finding, Zhang_2017_ICCV, Zhang18_unsupSaliency, Nguyen2019usps}.
Despite these efforts, the gap between unsupervised and fully supervised SOD methods remains significant.

\begin{figure}
    \centering
    \includegraphics[width=.48\textwidth]{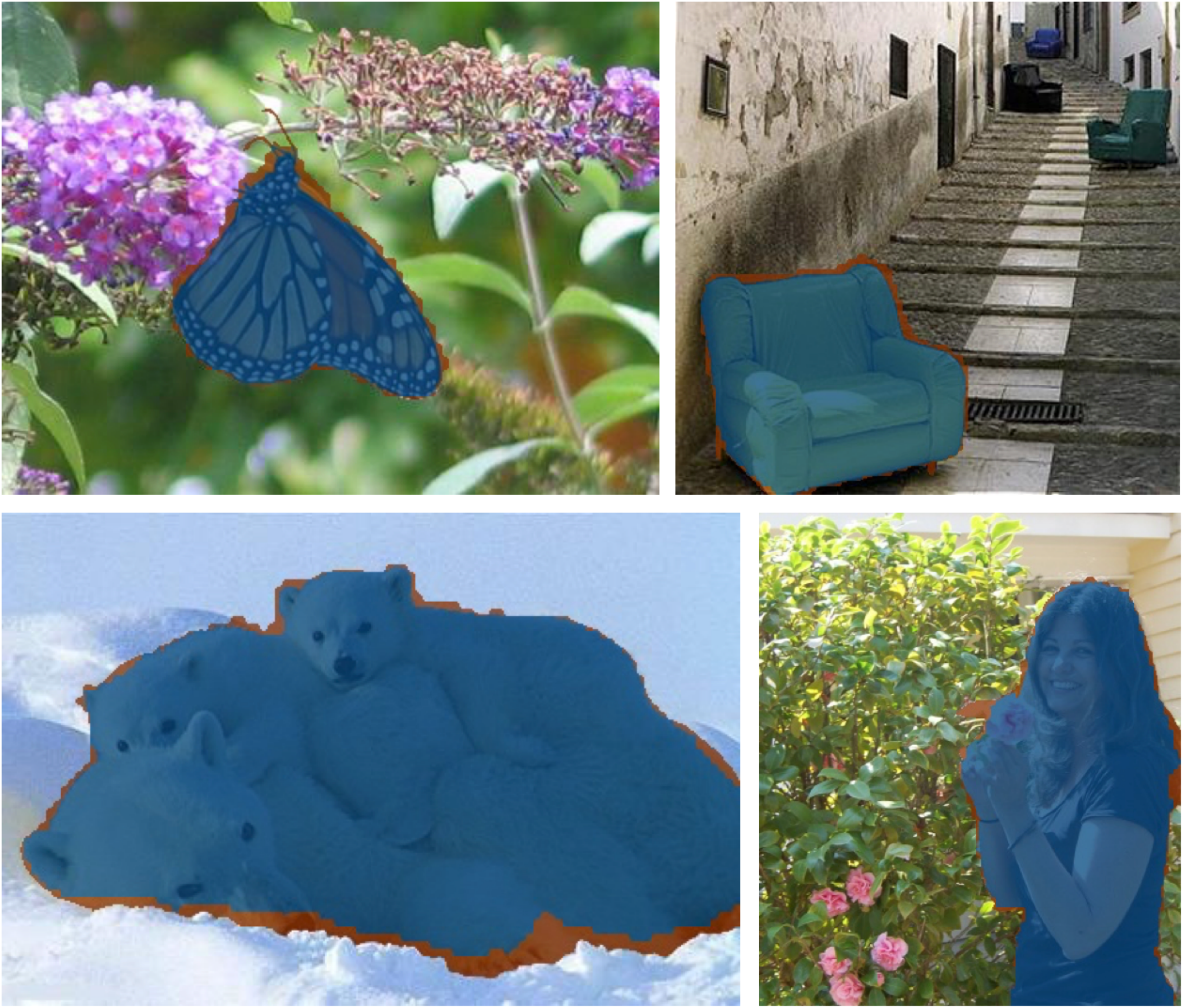}
    \caption{In this work, we propose \methodName, a framework for salient object detection that employs no human annotation.
    The figure depicts example segmentations produced by our model on DUT-OMRON~\cite{yang2013saliency}, DUTS-TE~\cite{wang2017learning} and ECSSD~\cite{shi2015hierarchical}, with blue and orange regions denoting the intersection and the difference between the predicted mask and ground-truth label, respectively.
    Despite no supervision, \methodName reliably segments a close approximation to the full spatial extent of salient regions. Best viewed in colour.
    }
    \vspace{-.5cm}
    \label{fig:teaser}
\end{figure}

Interestingly, however, it has been noted that recent self-supervised models such as DINO~\cite{Caron_2021_ICCV} exhibit significant potential for object segmentation despite the fact that their training objective does not explicitly encourage pixel grouping. 
The focus of this work is to leverage this observation to propose a \textit{simple yet effective mechanism for extracting object regions from self-supervised features} that can be employed for the task of unsupervised salient object detection.

To this end, we explore the use of spectral clustering~\cite{shi2000normalized}, 
a classical graph-theoretic clustering algorithm, 
and find that it can generate useful segmentation candidates across a range of self-supervised features (i.e., DINO~\cite{Caron_2021_ICCV}, MoCov2~\cite{he2020momentum}, and SwAV~\cite{Caron20}).
Motivated by this finding, 
we propose a simple \textit{winner-takes-all} voting strategy to select the salient object masks among a collection of clusters produced by repeated applications of spectral clustering to self-supervised features.\footnote{
In this work, we use the terms \textit{cluster} and \textit{mask} interchangeably. 
Specifically, by mask, we mean a (one-hot) mask which encodes the spatial extent of a cluster.}
In particular, we base our voting strategy on two priors: 
The first is a \tit{\framingPrior prior} that a salient object should not occupy the full image height or width;
The second is a \tit{\distinctivenessPrior prior} that assumes that salient regions are sufficiently distinctive that they will appear as clusters among an appropriately constructed collection of redundant re-clusterings of the data.
We then show that the selected salient masks can be employed as \textit{pseudo-labels} to train a saliency estimation network that achieves state-of-the-art results on a variety of benchmarks.

In summary, we make the following contributions:
(i)~We revisit spectral clustering and highlight its benefits over \kmeans clustering as a proposal mechanism to identify object regions from self-supervised features on three popular salient object detection (SOD) datasets; %
(ii)~We propose an effective voting strategy to select the most salient object mask in an image 
among multiple segmentations generated from different self-supervised features by leveraging saliency priors;
(iii)~Using the salient masks as pseudo ground truth masks (pseudo-masks), 
we train an object segmentation model, \methodName, that outperforms the previous unsupervised saliency detection approaches on three SOD benchmarks.

%% file: sections/02-related.tex
\section{Related work} \label{sec:related}

Our work relates to two themes in the literature: \textit{self-supervised representation learning} and \textit{unsupervised saliency detection}.
Each of these is discussed next.

\subsection{Self-supervised representation learning}

There has been a great deal of interest in self-supervised approaches to learning visual representations that obviate the requirement for labels. 
These include techniques for solving proxy tasks, 
such as predicting patch locations~\cite{doersch2015unsupervised},
patch discrimination~\cite{dosovitskiy2015discriminative},
grouping through co-occurrence~\cite{isola2015learning},
colourisation~\cite{larsson2017colorization},
jigsaw puzzles~\cite{noroozi2016unsupervised},
common fate principles~\cite{mahendran2018cross},
clustering~\cite{caron2018deep,Yuki20},
and instance discrimination~\cite{wu2018unsupervised,he2020momentum}. 
Our work is inspired by recent work exploring the use of
self-supervised transformers~\cite{chen2021empirical,Caron_2021_ICCV},
and by the analysis provided by~\cite{Caron_2021_ICCV,naseer2021intriguing}
who noted that the self-attention of Vision Transformers (ViT)~\cite{dosovitskiy2015discriminative} are capable of highlighting spatially coherent object regions in their input.
While prior work~\cite{zhan2018mix} has demonstrated that self-supervised pretraining can be effective
for semantic segmentation when coupled with end-to-end supervised metric learning on the target dataset, we instead seek a simple way to exploit self-supervised features for object segmentation without annotation.

\subsection{Unsupervised saliency detection}
Supervised object segmentation requires pixel-wise annotations which are time-consuming to acquire.
Seeking to avoid this cost, many attempts have been made to solve the task in an unsupervised fashion. Prior to the dominance of deep neural networks, a broad range of handcrafted methods were proposed~\cite{liu2007cvpr, Achanta2009cvpr, klein2011iccv, goferman2012tpami, yan2013cvpr, zhu2014cvpr, cheng2015tpami} based on one or more priors relating to foreground regions within an image such as the contrast prior~\cite{itti1998tpami}, 
centre prior~\cite{judd2009iccv}, and boundary prior~\cite{wei2012eccv}. However, these handcrafted approaches suffer from poor performance relative to recent DNN-based models, described next.

\vspace{5pt}
\noindent \textbf{Generative models.}
A common approach for DNN-based unsupervised object segmentation is to utilise generative adversarial networks (GANs)~\cite{goodfellow2014neurips}. Specifically, given an image, a generator is adversarially trained to produce an object mask which will be used to composite a realistic image by copying the corresponding object region in the image into a background which is either synthesised~\cite{bielski2019emergence} or taken from a different image~\cite{arandjelovic2019object}.
An alternative family of approaches aims to discover a direction in the latent space of a pre-trained GAN that can be used to segment foreground and background regions~\cite{voynov2020unsupervised, voynov2021obj, melas2021finding}.
Then, a saliency detector is trained on a synthetic data set composed of pairs of images together with their foreground masks generated via the discovered latent space structure.
In contrast, we seek to exploit representations learned via self-supervision by discriminative, rather than generative, models.

\vspace{5pt}
\noindent \textbf{Noisy supervision with pseudo-labels.}
More closely related to our work, the use of weak pseudo saliency masks for training a DNN has been proposed.
SBF~\cite{Zhang_2017_ICCV}, the first attempt to train  saliency detector without human annotation, proposed to train a model with superpixel-level pseudo-masks generated by fusing weak saliency maps from multiple unsupervised methods (\ie, ~\cite{zhang2015iccv, zhang2016tpami, shi2016tpami}).
Similarly, USD~\cite{Zhang18_unsupSaliency} aims to learn from diverse noisy pseudo-labels obtained via distinct unsupervised handcrafted methods in such a way that a saliency detector trained with the pseudo-masks can predict a saliency map free from label noise.
DeepUSPS~\cite{Nguyen2019usps} proposed to refine the pseudo-masks for images produced by handcrafted saliency methods, 
by training segmentation networks via a self-supervised iterative refinement process.
The refined pseudo-masks are then combined from different handcrafted methods to train a final segmentation network.
In contrast to the methods above,  we use neither handcrafted saliency methods nor an iterative refinement strategy, resulting in a simpler learning framework.

\vspace{5pt}
\noindent \textbf{Object segmentation properties of self-supervised vision transformers.}
Another line of related work has sought to investigate the observation that self-supervised ViTs~\cite{Caron_2021_ICCV} exhibit object segmentation potential.
LOST~\cite{Sim2021} propose to pick a seed patch from such a ViT that is likely to contain part of a foreground object, and then expand the seed patch to different patches sharing a high similarity with the seed patch.
Concurrent work, TokenCut~\cite{wang2022cvpr}, proposes to use Normalised Cuts~\cite{shi2000normalized} to segment the salient object among the final layer self-attention key features of ViT.
\methodName differs from the TokenCut approach to salient object detection in two key ways:
(1) While we similarly employ spectral clustering as part of our pipeline, we demonstrate the significant additional value of integrating cues from diverse re-clusterings via voting to bootstrap a pseudo-labelling process;
(2) Thanks to the flexibility of our clustering approach, we are able to leverage saliency cues from self-supervised convolutional neural networks (CNNs) as well as ViT architectures, and show the benefits of doing so.
We compare our approach with theirs in \cref{sec:experiments}.

%% file: sections/03-method.tex
\input{figures/overview}

\section{Method}
In this section, we begin by formalising the problem scenario (Sec.~\ref{subsec:problem}) and briefly summarise spectral clustering (Sec.~\ref{subsec:spectral}).
Then, we introduce our approach to address unsupervised salient object detection by selecting pseudo ground truth masks via spectral clustering, 
and train a saliency prediction network, called \methodName~(Sec.~\ref{subsec:inseg}).

\subsection{Problem formulation}
\label{subsec:problem}
Here, we consider the task of unsupervised salient object detection~(SOD),
with the goal of training a segmenter~($\mathrm{\Phi}_{\mrm{seg}}$) that seeks to partition the image into two disjoint groups, namely foreground and background. 
That is,
\begin{align}
    \mathrm{\Phi}_{\mrm{seg}}(\mathbf{I}; \mathrm{\Theta}) =
    \mathbf{M}_{\mathrm{seg}} \in \{0, 1\}^{H\times W}
\end{align}
where $\mathbf{I} \in \mathbb{R}^{H \times W \times 3}$ refers to an input image, 
and $\mbf{\Theta}$ represents learnable parameters.
$\mathbf{M}_{\mathrm{seg}}$ denotes a binary segmentation mask, with 1s denoting a foreground region and 0s denoting background.

Traditionally this has been treated as a clustering problem where the key challenge lies in designing effective features for accurately describing salient regions. 
In this work, we instead look for a simple yet effective solution by leveraging self-supervised visual representations.

\subsection{Segmentation with spectral clustering}
\label{subsec:spectral}
Conceptually, segmentation is obtained via spectral clustering~\cite{shi2000normalized} with pixel-wise image features by projecting the features onto a representation space such that image partitions can be decided by directly comparing similarities between their corresponding features.

Concretely, we first extract dense features from the image, 
{\em i.e.}, $\mathbf{F} \in \mathbb{R}^{h \times w \times D}$ with a pre-trained convolutional- or transformer-based encoder. 
Then, each feature vector $\mathbf{f}_i \in \mathbb{R}^{D}$ in the dense feature map $\mathbf{F}$ can be seen as a vertex in an undirected graph with vertex set $\mathbf{V} = \{\mathbf{f}_1, \dots, \mathbf{f}_N\} \in \mathbb{R}^{N \times D}$, where $N = h \cdot w$.
Each edge between two vertices $\mathbf{f}_i$ and $\mathbf{f}_j$ is associated with a non-negative weight $w_{ij} \geq 0$ defined by feature similarity. 
In particular, the weighted {\em adjacency matrix} $\mathbf{W}$ of the graph is computed as
\begin{align}
\mathbf{W} = (w_{ij}) = \frac{\mathbf{V}\mathbf{V}^T }{\|\mathbf{V}\| \|\mathbf{V}\|} \in \mathbb{R}^{N \times N}
\end{align}
\noindent 
where the \textit{degree} of a vertex $\mathbf{f}_i \in \mathbf{V}$ is defined as $d_i=\sum_{j=1}^{N} w_{ij}$,
and the {\em degree matrix} $\mathbf{D}$ is defined with the degrees $d_1, \dots, d_n$ on the diagonal. Given the adjacency matrix $\mathbf{W}$ and the degree matrix $\mathbf{D}$, the (un-normalised) graph Laplacian $\mathbf{L}$ is defined as:
\begin{align}
&\mathbf{L}  = \mathbf{D} - \mathbf{W}\label{eq}
\end{align}

\noindent Given $\mathbf{L}$, we can solve the generalised eigenproblem:
\begin{align}
\mathbf{Lu} = \lambda \mathbf{Du}
\end{align}
where $\mathbf{u} \in \mathbb{R}^{N}$ and $\lambda$ represent an eigenvector and its eigenvalue.
We take the $k$ eigenvectors with the lowest eigenvalues and form a matrix $\mathbf{U} \in \mathbb{R}^{N\times k}$ which has the eigenvectors as its columns.

Finally, a set of clusters $\mathcal{C}$ is obtained by running \kmeans algorithm on the row vectors of a matrix $\mathbf{U}$ (see supplementary for more detail), 
producing regions with all pixels from the corresponding cluster.
Note that, at this stage, the resulting clusters are composed of both object and background masks---the object mask itself will be selected by our selection strategy described next.

\subsection{Supervision with pseudo-mask from spectral clusters}
\label{subsec:inseg}
Here, we first introduce our voting strategy for selecting a salient mask from a set of spectral clusters from different features and multiple $k$ values (\ie, cluster numbers), which utilises a \tit{\framingPrior prior} and \tit{\distinctivenessPrior prior}.
Then, we describe our model, \methodName, which is trained by using the selected salient masks as pseudo-masks for supervision.

\subsubsection{Spectral cluster winner-takes-all voting}
\label{subsec:mask-selection}
To choose the salient object among the mixture of foreground and background masks generated by spectral clustering,  we propose a voting strategy based on two observations:
(1) The spatial extent of an object rarely occupies the entire height and width of an image.
(2) Salient object regions are likely to appear in multiple clusters across different self-supervised features as well as with different cluster numbers $k$.
In other words, among $k$ clusters from an application of clustering, we assume that at least one cluster encodes an object region within the image and that this holds for different features (\eg, DINO, MoCov2, or SwAV).
We call these priors the \tit{\framingPrior prior} and \tit{\distinctivenessPrior prior}, respectively.
Note that the \framingPrior prior bears a resemblance to the centre prior~\cite{judd2009iccv}, which states that salient objects are likely to be located near the center of an image, and the boundary prior~\cite{wei2012eccv}, which presumes that the foreground object rarely touches the boundary of an image. However, the \framingPrior prior differs from these priors in that it is not related to a location of an object but rather the scale of an object within an image.

To use the \framingPrior prior and \distinctivenessPrior prior in practice, we first form a candidate set of masks by repeatedly applying spectral clustering to different features with multiple $k$ values.
Then, we treat masks whose spatial extent is as long as the width or height of the image as background masks, and eliminate them from the candidate set.
Finally, we employ \textit{winner-takes-all} voting: we pick the mask with the highest average pair-wise similarity w.r.t.~IoU among all remaining masks as the final mask for salient objects (bottom of Fig.~\ref{fig:overview}).
There are two edge cases in the background elimination process we handle explicitly:
(i) when no masks are left in the candidate set and
(ii) when only two masks are left, sharing the same IoU.
For the former case, we simply keep all the masks in the candidate set as every mask highlights regions spanning the spatial extent of the image, breaking the assumption of the \framingPrior prior.
For the latter, we break ties randomly to pick one of the two masks.

\subsubsection{\methodName}\label{method:selfmask}
Here, we describe our model architecture, training, and inference procedure.

\vspace{5pt}
\noindent {\bf Architecture.}
We base our salient object detection network, called \methodName, on a variant of the MaskFormer architecture~\cite{cheng2021perpixel} which was originally proposed for the semantic/instance segmentation task.

\methodName has two different sets of outputs: mask predictions and objectness scores for each mask. 
In detail, our model comprises an image encoder, a pixel decoder, 
and a transformer decoder (upper part of Fig.~\ref{fig:overview}). 
The image encoder takes an image $\mathbf{I} \in \mathbb{R}^{H \times W \times 3}$ as input and outputs feature maps $f(\mathbf{I}) \in \mathbb{R}^{h \times w \times D}$.
Then, the feature maps are fed into the pixel decoder to produce upsampled features $g(f(\mathbf{I})) \in \mathbb{R}^{H \times W \times D}$.
The feature maps are also passed to the transformer decoder which outputs $n_q$ per-mask embeddings by using the feature maps as keys and values and the learnable embeddings as queries.
The final mask predictions $\mbf{M}$ are produced via matrix multiplication between the upsampled features and per-mask embeddings, followed by an element-wise sigmoid function $\sigma\p{\cdot}$:
\begin{equation}
    \begin{split}
    \mbf{M} = \{\mbf{M}_i~|~\mbf{M}_i = \sigma\p{g\p{f\p{\mbf{I}}} \mbf{q}_i}, 
    i=1, ..., n_q\}
    \end{split}
\end{equation}
where $\mbf{q}_i \in \mbb{R}^{D}$ denotes the $i$th query (\ie, per-mask embedding).
For each mask $\mbf{M}_i$, an objectness score $o_i \in [0, 1]$ is estimated by feeding the corresponding per-mask embedding $\mbf{q}_i$ to a simple MLP with two hidden layers followed by a sigmoid function. 
Note that other encoder and decoder architectures can also be used here.

\begin{figure*}[!t]
    \centering
    \includegraphics[width=\textwidth]{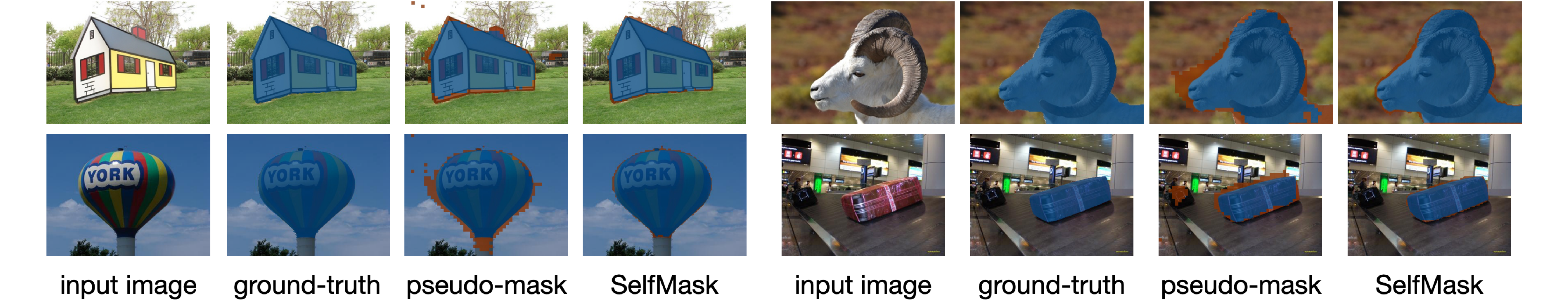}
    \caption{Sample visualisations of the pseudo-masks and predictions from our model on the ECSSD, DUT-OMRON, and DUTS-TE benchmarks. 
    From left to right, the input image, ground-truth mask, a pseudo-mask decided by the proposed voting-based salient mask selection, and the prediction of our model are shown.
    Blue and orange coloured regions denote the intersection and difference between a ground-truth and a predicted mask.
    Best viewed in colour.}
    \mbox{}\\
    \label{fig:visualisations}
\end{figure*}

\vspace{5pt}
\noindent {\bf Objective.} 
For training, we employ two objective functions: a mask loss and a ranking loss, denoted by $\mcal{L}_{\text{mask}}$ and $\mcal{L}_{\text{rank}}$, respectively. 
Given $n_q$ mask predictions for an image from the model, we encourage all predictions to be similar to the pseudo-mask.
Specifically, following~\cite{cheng2021perpixel}, we use the Dice coefficient~\cite{miletari2016v-net}, which considers the class-imbalance between foreground and background regions within an image, as the mask loss.
It is worth noting that, unlike~\cite{cheng2021perpixel}, we do not include the focal loss~\cite{Lin2017iccv} in the mask loss since we find that it hinders convergence.

To decide which prediction best highlights the salient region in the image among the proposed candidates when $n_q > 1$, we rank the predicted masks based on their objectness score.
Specifically, we first re-order the indices of the predicted masks by their mask loss in ascending order such that
\begin{equation}
    \mcal{L}_\text{mask}\p{\mbf{M}_i, \mbf{M}_\text{pseudo}} \leq \mcal{L}_\text{mask}\p{\mbf{M}_j, \mbf{M}_\text{pseudo}}
\end{equation}
for any $i < j$, where $\mbf{M}_\text{pseudo}$ denotes the target pseudo-mask for the image.
Then, we enforce the objectness score $o_i$ of the mask $\mbf{M}_i$ to be higher than the scores $o_j$ for any $j > i$.
As a consequence the model is encouraged to produce a higher score for a predicted mask that more closely resembles the pseudo-mask than other predictions.
We instantiate this ranking loss as a \textit{hinge loss}~\cite{cortes1995svm}: 
\begin{equation}
    \mathcal{L}_{\text{rank}} = \sum^{n_q - 1}_{i=1}\sum^{n_q}_{j > i}\max\p{0,~o_j - o_i}.
\end{equation}

\noindent Overall, our final objective function is as follows:
\begin{equation}
    \mcal{L} = \mcal{L}_{\text{mask}} + \lambda\mcal{L}_{\text{rank}}
\end{equation}
where $\lambda$ is a weighting factor, which is set to 1.0 across our experiments. 
Following~\cite{carion2020detr, cheng2021perpixel}, 
we compute the loss for outputs from each layer of the transformer decoder.

\vspace{5pt}
\noindent {\bf Inference.} 
During inference, 
given $n_q$ predicted masks for an image and their objectness score, 
we pick the mask with the highest score as the salient object detection and binarise it with a fixed threshold of 0.5.

%% file: figures/overview.tex
\begin{figure*}[t]
    \centering
    \includegraphics[width=\textwidth]{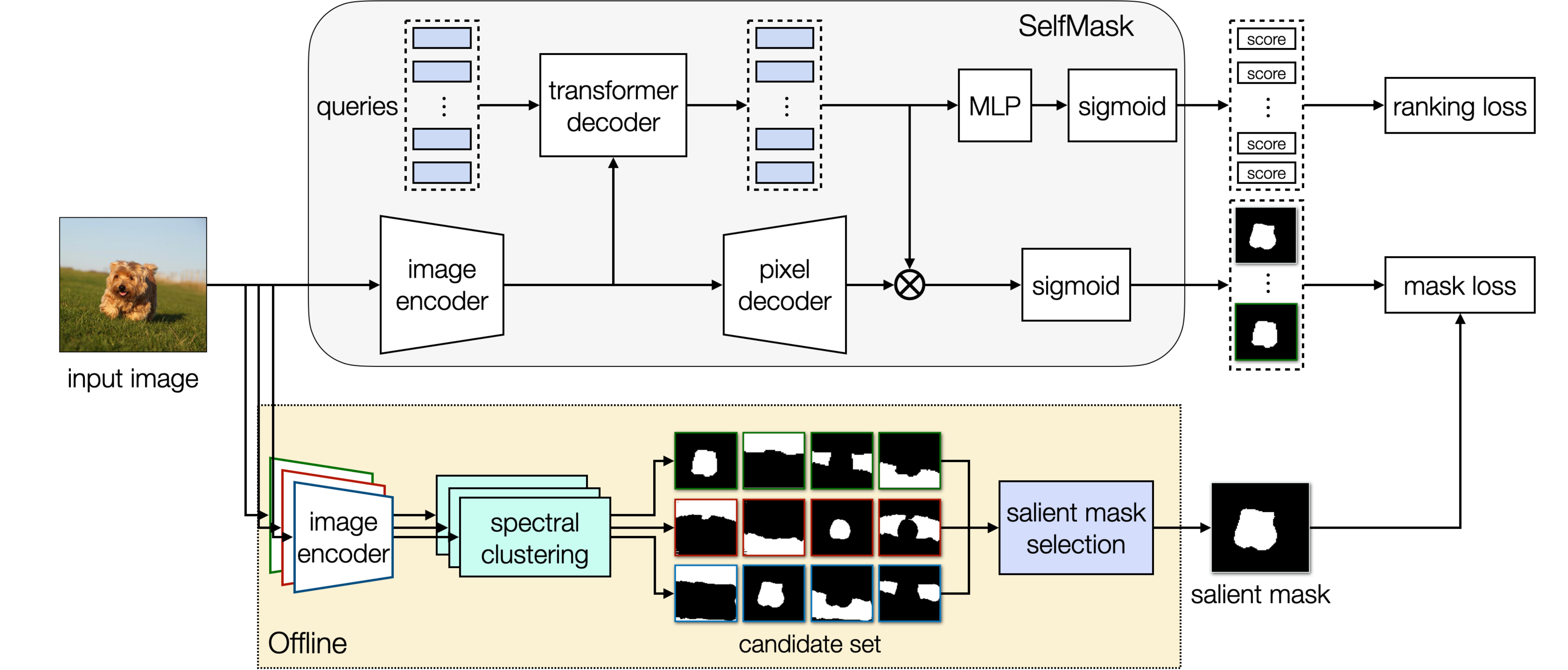}
    \vspace{-.3cm}
    \caption{Overview of our approach. Given different self-supervised encoders, we first generate a set of pseudo-mask candidates per image using spectral clustering before the training step. In the figure we show 12 masks from clusterings from three different encoder features with $k$=4. We select the most salient mask among them via the proposed voting strategy and use it as a pseudo-mask for the image.
    Then we train our model to predict $n_q$ queries (\ie, predictions), all of which are encouraged to be similar to the salient mask.
    To make the model aware of the objectness of each prediction, we use the ranking loss which encourages the objectness score of a prediction closer to the salient mask to be higher.
    At inference time, we select the prediction with the highest objectness score.
    Please see the text for details.}
    \mbox{}\vspace{-0.6cm}\\
    \label{fig:overview}
\end{figure*}

%% file: sections/04-experiments.tex
\section{Experiments} 
\label{sec:experiments}
In this section, 
we first describe the datasets used in our experiments (Sec.~\ref{exp:subsec:datasets})
and provide implementation details (Sec.~\ref{exp:subsec:impl}). 
We then conduct ablation study (Sec.~\ref{exp:ablation}) and report our results for salient object detection (Sec.~\ref{exp:comparison-os}).

\subsection{Datasets}
\label{exp:subsec:datasets}
We use \textit{DUTS-TR}~\cite{wang2017learning}, which contains 10,553 images, to train our model with the pseudo-masks generated by following Sec.~\ref{subsec:inseg}. 
We emphasize that only images are used for generating pseudo-masks and training, without the corresponding labels.
For our ablation study and comparison to previous work, we consider five popular saliency datasets including \textit{DUT-OMRON}~\cite{yang2013saliency}, 
which comprises 5,168 images of varied content with ground-truth pixel level masks;
\textit{DUTS-TE}~\cite{wang2017learning}, containing 5,019 images selected from the
SUN dataset~\cite{xiao2010sun} and ImageNet test set~\cite{deng2009imagenet};
\textit{ECSSD}~\cite{shi2015hierarchical} which contains 1,000 images that were selected to represent complex scenes; \textit{HKU-IS}~\cite{Li_2015_CVPR} which consists of 4,447 scene images with foreground/background sharing the similar appearances; \textit{SOD}~\cite{Movahedi2010} which contains 300 images with many images having multiple salient objects.

\subsection{Implementation details}
\label{exp:subsec:impl}

{\noindent \bf Networks. }
We use the ViT-S/8 architecture~\cite{dosovitskiy2021vit} for the encoder, a bilinear upsampler with a scale factor of 2 for the pixel decoder, and 6 transformer layers~\cite{vaswani2017transformer} for the transformer decoder. 
For the MLP applied to per-mask queries (with a dimensionality of 384) that outputs a scalar value for the objectness score, we use three fully-connected layers with a ReLU activation between them.
We set the same number of units for the hidden nodes as the input (\ie, 384) and output a single value followed by a sigmoid.

\vspace{5pt}
{\noindent \bf Training details. } 
We train our models for 12 epochs and optimise all parameters including the backbone encoder using AdamW~\cite{loshchilov2018iclr} with a learning rate of 6e-6 and the Poly learning rate policy~\cite{Ouali_2020_CVPR, Xie2020DEALDA, Chen_2018_ECCV}. 
For data augmentation, we use random scaling with a scale range of [0.1, 1.0], random cropping to a size of 224$\times$224 and random horizontal flipping with a probability of 0.5. 
In addition, photometric transformations include random color jittering, 
color dropping, and Gaussian blurring are applied.
We run each model with three different seeds and report the average.

\vspace{5pt}
{\noindent \bf Metrics. } 
In our experiments,  we report intersection-over-union (IoU), pixel accuracy (Acc) and maximal $F_\beta$ score (max $F_\beta$) with $\beta^2$ set to 0.3  following~\cite{wang2022cvpr}.
Please refer to the supplementary for more details on these metrics.

\subsection{Ablation study}
\label{exp:ablation}
In this section, 
we first conduct experiments to compare the effectiveness of spectral clustering and \kmeans when applied to self-supervised image encoders.
Next, we quantitatively verify the performance of our winner-takes-all voting strategy for foreground mask selection and compare it to different saliency selection methods. 
Lastly, we investigate the effect of different number of queries on \methodName.

\input{tables/k-means-vs-spectral-self}
\input{tables/k-means-vs-spectral-fully-sup}

\subsubsection{Spectral clustering vs $k$-means clustering} 
We compare spectral clustering against a \kmeans~\cite{lloyd1982least} clustering baseline on three salient object detection benchmarks. 
As the resulting segmentations from each algorithm are agnostic to foreground/background regions,  
we consider a best-case evaluation for both algorithms.
In detail, given a groundtruth mask, we pick the cluster with the highest IoU w.r.t. the groundtruth. Such IoUs act as an upper-bound score among the clusters (\ie, average best overlap in~\cite{APBMM2014}). 

To account for the effect of the cluster number $k$ on performance of the clustering algorithms, we consider different $k$ values from 2 to 4 and average the results. For the full results with each $k$ value, please refer to the supplementary.

As shown in Tab.~\ref{tab:kmeans-vs-spectral-self},
we observe that object masks from spectral clustering consistently outperform \kmeans masks by a large margin.
Interestingly, however, when using fully-supervised image encoders, 
the performance gain of spectral clustering diminishes (Tab.~\ref{tab:kmeans-vs-spectral-fully-sup}).
These findings boil down to a simple summary: 
\textit{while using \underline{self-supervised} visual representations for grouping, spectral clustering is considerably superior to \kmeans, 
regardless of the choice of encoder architecture and self-supervised learning algorithm.}

\subsubsection{Voting for salient object masks}
Here, we conduct experiments to assess our voting method for selecting a foreground mask among the mask candidates.
Since these experiments include ablations across hyperparameter choices, 
we conduct them on the HKU-IS~\cite{Li_2015_CVPR} and SOD~\cite{Movahedi2010}, rather than the benchmarks used to compare to prior work.

In detail, we construct an initial corpus of mask candidates by  clustering different self-supervised features with different number of clusters, as described in Sec.~\ref{subsec:spectral}.
For this, we build the candidate set with spectral clusters from different combinations of self-supervised features, \eg, MoCov2/DINO or SwAV/MoCov2.
It is important to do so to allow our voting-based method to leverage the \distinctivenessPrior prior across different features.
In addition, for each combination, we experiment with 3 different $k$ value settings: $k$=$2$, $\{2, 3\}$ or $\{2,3,4\}$ to account for various object scales, 
\eg,~a lower $k$ tends to cover large regions, 
while a higher $k$ segments smaller objects. 
We then evaluate the selected masks on the \textit{HKU-IS} set.
For reference we also compute
an upper bound IoU, which is computed in a similar way as done in the previous section.

\vspace{5pt}
{\noindent \bf Effectiveness of clustering various self-supervised models with different number of clusters. }
As shown in Tab.~\ref{tab:effect-of-feature-combinations}, 
we make two observations:
(i) IoU of both selected masks and upper bound masks, denoted by pseudo-mask and UB
improves by increasing $k$ across all feature combinations;
(ii) using all three features (\ie, DINO, MoCov2 and SwAV) results in better pseudo-masks than using two of the three features (\eg, MoCov2 and SwAV).
These support the \distinctivenessPrior prior, 
which assumes that at least one cluster represents a foreground region, and its application to our voting-based saliency selection method.

\vspace{5pt}
{\noindent \bf Effectiveness of the proposed voting approaches. }
We further validate the effectiveness of our voting scheme by comparing to different selection methods, \ie, random selection and a centre prior~\cite{judd2009iccv}-based strategy.
Specifically, we form a candidate set using DINO/MoCov2/SwAV features with $k=\{2, 3, 4\}$, which amounts to 27 masks in total.
For the random strategy, 
we simply pick one of the masks uniformly from the candidates.
For the centre prior selection strategy, 
we choose a mask whose average Euclidean distance to the image centre from its constituent pixel locations is lowest.
In addition, we consider each method with or without utilising the \framingPrior prior to assess the influence of filtering out background mask.
We evaluate the IoU of each case on the \tit{HKU-IS} and \tit{SOD} benchmarks.

As can be seen in Tab.~\ref{tab:comp-selection-strategies}, 
deploying the \framingPrior prior boosts IoU in all considered selection methods, with the proposed voting selection method performing best.
The \framingPrior prior plays a crucial role in the voting process: voting without this prior performs much worse than its counterparts in both random and center-based selections on \tit{HKU-IS}, 
and performs similarly to the random strategy on \tit{SOD}.
This is caused in large part by mistakes when selecting background masks as salient objects.

\input{tables/effect-of-feature-combinations}

\input{tables/comp-selection-strategies}

\input{figures/effect-of-n-queries}

\input{tables/os-comp}

\subsubsection{The influence of the number of queries} 
As described in Sec.~\ref{subsec:inseg},  we train \methodName using the selected salient masks as pseudo-masks. 
Here, 
we investigate the effect of the number of queries $n_q$ in the Transformer decoder.
For this, we train our model with $n_q$=\{5, 10, 20, 50, 100\} on \textit{DUTS-TR} and evaluate performance on the \textit{HKU-IS} benchmark in terms of max$F_\beta$ for two settings,
(i) using ground-truth masks to pick the best mask out of $n_q$ mask predictions, denoted as the SelfMask upper bound~(UB);
(ii) taking the mask with the highest object score, denoted SelfMask.

As shown in Figure~\ref{fig:ablation}, 
both SelfMask and SelfMask UB are fairly robust to the number of queries, initially increasing slightly with this hyperparameter (\ie, predictions)
before degrading after 20 queries.
We conjecture that this is because a handful of queries are enough to localise the salient objects, while further predictions may make it challenging to appropriately rank the objectness of each prediction. 
For this reason, in the section that follows, we consider \methodName with 20 queries, and pick the query with the highest objectness as our prediction during inference. 

\subsection{Comparison to state-of-the-art unsupervised saliency detection methods}
\label{exp:comparison-os}
To compare with existing works on unsupervised SOD,
we evaluate on three popular SOD benchmarks in terms of Acc., IoU, and max$F_\beta$. 
Following~\cite{wang2022cvpr}, we also report results after post-processing predictions with the bilateral solver~\cite{barron2016eccv}. 
As shown in Tab.~\ref{tab:os-comp}, 
while the pseudo-masks from spectral cluster voting already perform reasonably well compared to previous models, our self-trained model outperforms all existing approaches on all benchmarks.
This suggests both that the model can learn to generalise effectively from noisy masks, and that the objectness score trained with the ranking loss is effective for picking the best salient mask.

\subsection{Broader Impact}
This work contributes a new framework to deliver performant unsupervised salient object detection. 
As such, it offers the potential to underpin a range of societally beneficial applications that are bottlenecked by annotation costs.
These include improved low-cost medical image segmentation, crop measurement from aerial imagery, and wildlife monitoring.
However, low-cost segmentation is a powerful dual-use technology, and we caution against its deployment as a tool for unlawful surveillance and oppression.

%% file: tables/k-means-vs-spectral-self.tex
\setlength\dashlinedash{0.5pt}
\setlength\dashlinegap{1.5pt}
\setlength\arrayrulewidth{0.3pt}
\setlength{\tabcolsep}{.75pt}  %

\begin{table}[!t]
\footnotesize
  \centering
        \begin{tabular}{ccc c c c}  %
        \toprule
        \multirow{2}{*}{\textbf{Model}}
        &\multirow{2}{*}{\textbf{Arch.}}
        &\multirow{2}{*}{\textbf{Cluster.}}
        &\multicolumn{1}{c}{\textbf{DUT-OMRON}}
        &\multicolumn{1}{c}{\textbf{DUTS-TE}}
        &\multicolumn{1}{c}{\textbf{ECSSD}}\\
        &
        &
        &$k$=$\{2, 3, 4\}$
        &$k$=$\{2, 3, 4\}$
        &$k$=$\{2, 3, 4\}$\\
        \midrule
        \multicolumn{6}{c}{\textbf{Convolutional Nets}}\\
        \midrule
        MoCov2~\cite{he2020momentum}
        & ResNet50
        & \kmeans
        &.375
        &.415
        &.500\\
        MoCov2~\cite{he2020momentum} & ResNet50
        & spectral
        &\textbf{.387}
        &\textbf{.454}
        &\textbf{.627}\\ \midrule
    
        SwAV~\cite{Caron20}
        & ResNet50
        & \kmeans
        & .399
        & .444
        & .542\\
        
        SwAV~\cite{Caron20}
        & ResNet50
        & spectral
        &\textbf{.401}
        &\textbf{.458}
        &\textbf{.590}\\ \midrule

        \multicolumn{6}{c}{\textbf{Vision Transformer}}\\
        \midrule
        DINO~\cite{Caron_2021_ICCV}
        &ViT-S/16
        &\kmeans
        &.377
        &.392
        &.541\\
        
        DINO~\cite{Caron_2021_ICCV}
        &ViT-S/16
        &spectral
        &\textbf{.394}
        &\textbf{.417}
        &\textbf{.577}\\ \midrule
        
        DINO~\cite{Caron_2021_ICCV}
        &ViT-S/8
        &\kmeans
        &.369
        &.377
        &.551\\
        
        DINO~\cite{Caron_2021_ICCV}
        &ViT-S/8
        &spectral
        &\textbf{.398}
        &\textbf{.411}
        &\textbf{.587}\\
        \bottomrule
        \end{tabular}
\caption{\textbf{Spectral clustering dominates \kmeans for self-supervised features.} We report upper bound IoUs to compare the quality of masks produced by \kmeans and spectral clustering on \textit{self}-supervised features with two different encoder architectures (i.e., convolution- and transformer-based encoder). We report the average of the results from $k$=$\{2, 3, 4\}$. \vspace{-.2cm}}
\label{tab:kmeans-vs-spectral-self}
\end{table}

%% file: tables/k-means-vs-spectral-fully-sup.tex
\setlength\dashlinedash{0.5pt}
\setlength\dashlinegap{1.5pt}
\setlength\arrayrulewidth{0.3pt}
\setlength{\tabcolsep}{.75pt}  %

\begin{table}[!t]
\footnotesize
  \centering
        \begin{tabular}{ccc c c c}  %
        \toprule
        \multirow{2}{*}{\textbf{Model}}
        &\multirow{2}{*}{\textbf{Arch.}}
        &\multirow{2}{*}{\textbf{Cluster.}}
        &\multicolumn{1}{c}{\textbf{DUT-OMRON}}
        &\multicolumn{1}{c}{\textbf{DUTS-TE}}
        &\multicolumn{1}{c}{\textbf{ECSSD}}\\
        &
        &
        &$k$=$\{2, 3, 4\}$
        &$k$=$\{2, 3, 4\}$
        &$k$=$\{2, 3, 4\}$\\
        \midrule
        \multicolumn{6}{c}{\textbf{Convolutional Nets}}\\
        \midrule
        ResNet~\cite{he2020momentum}
        & ResNet50
        & \kmeans
        &\textbf{.337}
        &\textbf{.354}
        &\textbf{.444}\\
        
        ResNet~\cite{he2020momentum} & ResNet50
        & spectral
        &.310
        &.327
        &.437\\ \midrule

        \multicolumn{6}{c}{\textbf{Vision Transformer}}\\
        \midrule
        ViT~\cite{dosovitskiy2021vit}
        &ViT-S/16
        &\kmeans
        &\textbf{.394}
        &\textbf{.411}
        &.542\\
        
        ViT~\cite{dosovitskiy2021vit}
        &ViT-S/16
        &spectral
        &.380
        &.400
        &\textbf{.551}\\
        \bottomrule
        \end{tabular}
\caption{\textbf{Spectral clustering and \kmeans perform comparably for fully-supervised features.}
We report upper bound IoUs to provide a comparison of mask quality comparison between \kmeans and spectral clustering applied to \textit{fully}-supervised features.
We report the average of the results from $k$=$\{2, 3, 4\}$. \vspace{-0.2cm}}
\label{tab:kmeans-vs-spectral-fully-sup}
\end{table}

%% file: tables/effect-of-feature-combinations.tex
\begin{table}[t]
\small
  \centering
        \begin{tabular}{ccc c c c c}
        \toprule
        \multicolumn{3}{c}{Features}&
        \multirow{2}{*}{$k$}&
        \multirow{2}{*}{Pseudo-mask}&
        \multirow{2}{*}{UB}\\ \cmidrule{1-3}
        DINO~\cite{Caron_2021_ICCV}&
        MoCov2~\cite{he2020momentum}&
        SwAV~\cite{Caron20}&
        &
        \\
        \midrule
        \multirow{3}{*}{\xmark}&\multirow{3}{*}{\checkmark}&\multirow{3}{*}{\checkmark}&2&.508&.562\\
        &&&2, 3&.561&.626\\
        &&&2, 3, 4&.580&.658\\
        \midrule
        \multirow{3}{*}{\checkmark}&\multirow{3}{*}{\xmark}&\multirow{3}{*}{\checkmark}&2&.473&.553\\
        &&&2, 3& .538& .644\\
        &&&2, 3, 4&.559&.682\\
        \midrule
        \multirow{3}{*}{\checkmark}&\multirow{3}{*}{\checkmark}&\multirow{3}{*}{\xmark}&2&.459&.546\\
        &&&2, 3&.536&.648\\
        &&&2, 3, 4&.566&.688\\
        \midrule
        \multirow{3}{*}{\checkmark}&\multirow{3}{*}{\checkmark}&\multirow{3}{*}{\checkmark}&2&.511&.584\\
        &&&2, 3&.567&.664\\
        &&&2, 3, 4&\textbf{.590}&\textbf{.698}\\
        \bottomrule
        \end{tabular}
\caption{\textbf{Forming a candidate set with various self-supervised features and multiple $k$ values improves IoU of both pseudo-masks and upper bound masks (UB).}
We compare cases with different combinations of self-supervised features and cluster numbers of $k$=2, $\{2, 3\}$ or $\{2,3,4\}$ on the HKU-IS benchmark.}
\label{tab:effect-of-feature-combinations}
\end{table}

%% file: tables/comp-selection-strategies.tex
\setlength{\tabcolsep}{2pt}
\begin{table}[t]
    \small
    \centering
    \begin{tabular}{c c c c}
    \toprule
    \multirow{1}{*}{Selection}&
    \multirow{1}{*}{Framing prior}&
    \multirow{1}{*}{HKU-IS~\cite{Li_2015_CVPR}}&
    \multirow{1}{*}{SOD~\cite{Movahedi2010}}\\
    \midrule
    \multirow{2}{*}{random}&
    \xmark&
    .206&
    .197
    \\
    &
    \checkmark&
    .464&
    .277
    \\
    \midrule
    \multirow{2}{*}{center}&
    \xmark&
    .362&
    .122
    \\
    &
    \checkmark&
    .442&
    .392
    \\
    \midrule
    \multirow{2}{*}{voting (ours)}&
    \xmark&
    .081&
    .200
    \\
    &
    \checkmark&
    \textbf{.590}&
    \textbf{.447}
    \\
    \bottomrule
    \end{tabular}
\caption{\textbf{Winner-takes-all voting and the \framingPrior prior both significantly improve mask quality.} 
We compare our voting strategy to different selection strategies along with the effect of \framingPrior prior under the IoU metric.
Selection is performed from a candidate set including DINO, MoCov2 and SwAV features with $k = \{2,3,4\}$.\vspace{-0.3cm}}
\label{tab:comp-selection-strategies}
\end{table}

%% file: figures/effect-of-n-queries.tex
\begin{figure}[t]
\centering
\footnotesize
\begin{minipage}[t]{0.9\linewidth}
\centering
  \includegraphics[width=\linewidth]{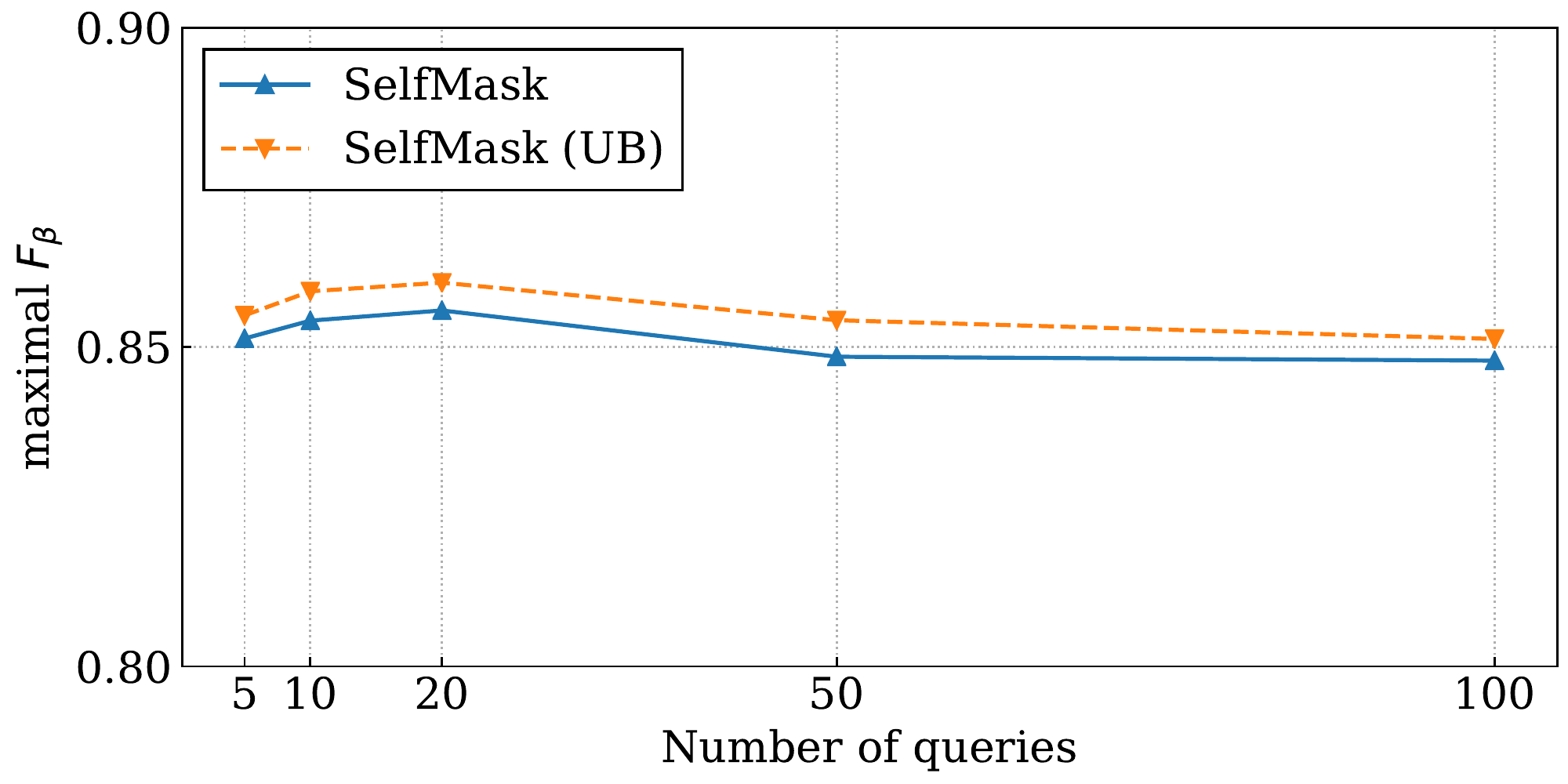}  
\end{minipage}
\vspace{-5pt}
\caption{{\bf Effect of number of queries on the performance of SelfMask on the HKU-IS dataset.} 
The model's prediction and its upper bound, denoted by SelfMask and SelfMask (UB) each, are shown.\vspace{-.2cm}}
\label{fig:ablation}
\end{figure}

%% file: tables/os-comp.tex
\setlength\dashlinedash{0.5pt}
\setlength\dashlinegap{3pt}
\setlength\arrayrulewidth{0.3pt}
\setlength{\tabcolsep}{10pt}

\setlength{\aboverulesep}{0pt}
\setlength{\belowrulesep}{0pt}
\begin{table*}[!t]
\small
  \centering
        \begin{tabular}{l 
        ccc@{\hspace{.5cm}}ccc@{\hspace{.5cm}}ccc}
        \toprule
        \multirow{2}{*}{Model}
         &\multicolumn{3}{c}{DUT-OMRON~\cite{yang2013saliency}}
         &\multicolumn{3}{c}{DUTS-TE~\cite{wang2017learning}}
         &\multicolumn{3}{c}{ECSSD~\cite{shi2015hierarchical}}\\   \cmidrule{2-10}
         &\small Acc &\scriptsize IoU&\scriptsize max$F_\beta$
         &\small Acc &\scriptsize IoU&\scriptsize max$F_\beta$
         &\small Acc &\scriptsize IoU&\scriptsize max$F_\beta$\\
        \midrule
        HS~\cite{yan2013cvpr}
        &.843&.433&.561
        &.826&.369&.504
        &.847&.508&.673\\
        wCtr~\cite{zhu2014cvpr}
        &.838&.416&.541
        &.835&.392&.522
        &.862&.517&.684\\
        WSC~\cite{li2015cvpr}
        &.865&.387&.523
        &.862&.384&.528
        &.852&.498&.683\\
        DeepUSPS~\cite{Nguyen2019usps}
        &.779&.305&.414
        &.773&.305&.425
        &.795&.440&.584\\
        BigBiGAN~\cite{voynov2021obj}
        &.856&.453&.549
        &.878&.498&.608
        &.899&.672&.782\\
        E-BigBiGAN~\cite{voynov2021obj}
        &.860&.464&.563
        &.882&.511&.624
        &.906&.684&.797\\
        Melas-Kyriazi et al.~\cite{melas2021finding}
        & .883 & .509 & -
        & .893 & .528 & - 
        &.915 & .713 & -\\
        LOST~\cite{Sim2021}
        &.797&.410&.473
        &.871&.518&.611
        &.895&.654&.758\\
        LOST~\cite{Sim2021} + Bilateral solver~\cite{barron2016eccv}
        &.818&.489&.578
        &.887&.572&.697
        &.916&.723&.837\\
        TokenCut~\cite{wang2022cvpr}
        &.880&.533&.600
        &.903&.576&.672
        &.918&.712&.803\\
        TokenCut~\cite{wang2022cvpr} + Bilateral solver~\cite{barron2016eccv}
        &.897&.618&.697
        &.914&.624&.755
        &.934&.772&.874\\
        \midrule
        pseudo-masks (Ours)
        &.811&.403&-
        &.845&.466&-
        &.893&.646&-
        \\
        \methodName (Ours)
        &.901&.582&.680
        &.923&.626&.750
        &.944&.781&.889\\
        \methodName (Ours) + Bilateral solver~\cite{barron2016eccv}
        &\textbf{.919}&\textbf{.655}&\textbf{.852}
        &\textbf{.933}&\textbf{.660}&\textbf{.882}
        &\textbf{.955}&\textbf{.818}&\textbf{.956}\\
        \bottomrule
        \end{tabular}
\vspace{-3pt}
\caption{ {\bf Comparison to the state-of-the-art unsupervised saliency detection methods on 3 salient object detection benchmarks.}
For all metrics, higher number indicates better results. The best score per column is highlighted in bold.
We observe that \methodName yields improved performance over prior state-of-the-art approaches across all benchmarks.
}
\label{tab:os-comp}
\end{table*}

%% file: sections/05-conclusion.tex
\section{Conclusion}

In this work, we address the challenging problem of unsupervised salient object detection (SOD).
For this, we first observe that self-supervised features exhibit significantly greater object segmentation potential with spectral clustering than with \kmeans.
Inspired by this observation, 
we extract foreground regions among multiple masks generated with multiple types of features,
and varying cluster numbers based on winner-takes-all voting.
By using the selected masks as pseudo-masks, 
we train a saliency detection network and show promising results compared to previous unsupervised methods on various SOD benchmarks.

%% file: sections/07-acknowledgement.tex
\vspace{5mm}

\noindent \textbf{Acknowledgements.}  
GS is supported by AI Factory, Inc. in Korea. 
WX is supported by Visual AI (EP/T028572/1). 
SA would like to thank Z. Novak and N. Novak for enabling his contribution.
GS would like to thank Jaesung Huh for proof-reading. 

%% file: sections/06-supp-mat.tex
\begin{appendices}

\appendix
In this supplementary material, 
we first describe the algorithm for spectral clustering (Sec.~\ref{supp:spectral-clustering}). 
Then, we briefly review the overall structures of the convolution- and transformer-based image encoders and how we extract dense features to which the spectral clustering is applied 
from each type of encoder (Sec.~\ref{supp:visual-encoder}).
The evaluation metrics are described in Sec.~\ref{supp:metrics}, and the full results for the comparison between \kmeans and spectral clustering with different cluster numbers, 
\ie, $k=\{2, 3, 4\}$ on the three main saliency benchmarks are shown in~Sec.~\ref{supp:kmeans-vs-spectral}.
Lastly, we describe typical failure cases of our model in Sec.~\ref{supp:visualsations}.

\section{Normalised spectral clustering algorithm}
\label{supp:spectral-clustering}
Here, we describe the normalised spectral clustering algorithm used to generate pseudo-masks for our model in Alg.~\ref{alg:spectral}.
\begin{algorithm}[!htb]
\small
\begin{minipage}{0.98\linewidth}
\caption{\textbf{ALGORITHM 1} Normalised spectral clustering~\cite{luxburg2004spectral,shi2000normalized}}
\label{alg:spectral}
\textbf{Input:} An adjacency matrix $\mathbf{W}\in\mathbb{R}^{N \times N}$, the number of clusters $k$ to be constructed.
\begin{algorithmic}[1]
    \STATE Compute the degree matrix $\mathbf{D}$ with $\mathbf{W}$.
    \STATE Compute the unnormalised Laplacian $\mathbf{L}$ using $\mathbf{W}$ and $\mathbf{D}$ using Eqn. 3.
    \STATE Compute the first $k$ generalised eigenvectors $\mathbf{u}_1, \dots, \mathbf{u}_k$ of the generalised eigen problem $\mathbf{Lu} = \lambda \mathbf{Du}$.
    \STATE Let $\mathbf{U} \in \mathbb{R}^{N \times k}$ be the matrix containing the vectors $\mathbf{u}_1, \dots, \mathbf{u}_k$ as the columns.
    \STATE For $i = 1, \dots, N$, let $\mathbf{y}_i \in \mathbb{R}^{k}$ be the vector corresponding to the $i$-th row of $\mathbf{U}$.
    \STATE Cluster the vectors $\{\mathbf{y}_i~|~i = 1, ..., N\} \in \mathbb{R}^{N \times k}$ with \kmeans into 
    clusters $\mathcal{C}_1, \dots, \mathcal{C}_k$.
\end{algorithmic}
{\bf Output:} Clusters $\mathcal{C}=\{\mathcal{C}_1, \dots, \mathcal{C}_k\}$
\end{minipage}
\end{algorithm}

Note that the adjacency matrix $\mbf{W}$ is computed using Eqn.~2 of the main paper, given the dense features from a visual encoder described next.

\input{tables/k-means-vs-spectral-full-results}

\section{Visual encoder}
\label{supp:visual-encoder}
Our approach utilises image representations learned by either convolution-based or transformer-based architectures to which spectral clustering will be applied. 
Here, we first briefly review how these feature representations are computed with each model.

\subsection{Convolution-based visual encoder} 
Convolutional neural networks (CNNs) for image representations, denoted by $\mrm{\Phi_{\text{CNN}}}$, 
consist of a series of 2D convolutional layers 
and non-linear activation functions which operate on an image in a sliding window fashion. 
Specifically, given an image $\mathbf{I} \in \mathbb{R}^{H \times W \times 3}$, 
the CNNs outputs dense feature maps $\mathbf{F}_{\text{CNN}} \in \mathbb{R}^{h \times w \times D}$ where $h$=$\frac{H}{s}$ and $w$=$\frac{W}{s}$ with $s$ denoting the total stride of the network and $D$ denotes the dimensionality of the features. That is,
\begin{align}
    \mathbf{F}_{\textsc{CNN}} = \mathrm{\Phi}_{\textsc{CNN}}(\mathbf{I})  \in \mathbb{R}^{h \times w \times D}
\end{align} 
where the parameters for the CNNs are omitted for simplicity.

\subsection{Transformer-based visual encoder} 
In the recent literature, 
Transformer-based architectures have shown tremendous success in the computer vision community, including ViT~\cite{dosovitskiy2021vit}, DeiT~\cite{touvron2021deit}, T2T-ViT~\cite{Yuan_2021_ICCV}, and BEiT\cite{beit}.
Generally speaking, these architectures consist of three components, 
namely, tokeniser~($\mathrm{\Phi}_{\textsc{TK}}$), linear projection~($\mathrm{\Phi}_{\textsc{LP}}$), and transformer encoder~($\mathrm{\Phi}_{\text{TE}}$):
\begin{align}
\mathbf{F}_{\text{Transformer}} = \mathrm{\Phi}_{\text{TE}} \circ \mathrm{\Phi}_{\textsc{LP}} \circ \mathrm{\Phi}_{\textsc{TK}}(\mathbf{I}) \in \mbb{R}^{h \times w \times D}
\end{align}
where $\mathbf{F}_{\text{Transformer}}$ denotes the dense features from a transformer-based encoder.

\vspace{5pt}
\noindent {\bf Tokeniser.}
Given an image as input, {\em i.e.}~$\mathbf{I} \in \mathbb{R}^{H \times W \times 3}$, the image is first divided by a tokeniser into non-overlapping patches of a fixed size $P \times P$,
ending up $N$ patches per frame, {\em i.e.}~$N = \frac{HW}{P^2}$:
\begin{align}
    \mathrm{\Phi}_{\textsc{TK}}(\mathbf{I}) =
    \{\mbf{x}_i~|~\mbf{x}_i = \mathrm{\Phi}_{\textsc{TK}}(\mathbf{I})_i \in \mathbb{R}^{3P^2},~i=1,...,N\}
\end{align}
where $\mbf{x}_i$ denotes the $i$th patch.
\\

\noindent {\bf Linear projection. }
Once tokenised, 
each patch from the image is fed through a linear layer $\mathrm{\Phi}_{\textsc{LP}}$ and projected into a vector (a.k.a. token):
\begin{align*}
\mathbf{z}_{i} = \mathrm{\Phi}_{\textsc{LP}}(\mathbf{x}_{i}) + \textsc{pe}_i \in \mathbb{R}^{D}
\end{align*}
where $\mathbf{x}_{i} \in \mathbb{R}^{3P^2}$ refers to the $i$th patch, 
and its corresponding learnable positional embeddings~$\textsc{PE}_i \in \mathbb{R}^{D}$ are added to the patch token $\mathrm{\Phi}_{\textsc{LP}}(\mathbf{x}_{i}) \in \mathbb{R}^{D}$. 
Then, the $N$ augmented patch tokens are concatenated altogether with a \textit{class token} {\fontfamily{qcr}\selectfont[CLS]}$\in \mathbb{R}^{D}$, producing the final input form of $\mathbb{R}^{\left(N + 1\right) \times D}$ to a sequence of transformer layers, described in the following.
\\

\noindent {\bf Transformer encoder.}
A transformer encoder is composed of multiple transformer layers, each of which is subdivided into a self-attention layer and multi-layer perceptrons (MLPs). The self-attention layer contains three learnable linear layers, each of which takes the input tokens and outputs either key $K$, value $V$, or query $Q$ of the same dimensionality as the input tokens, i.e., $\mathbb{R}^{\left(N + 1\right) \times D}$.\footnote{In practice, we use a single linear layer which maps the $D$ dimension of the input tokens to $3\times D$ and equally splits them into $Q$, $K$, and $V$.} Then, the self-attention layer outputs {\fontfamily{qcr}\selectfont softmax}($\frac{QK^T}{\sqrt{D}}$, {\fontfamily{qcr}\selectfont dim=-1})$V\in \mathbb{R}^{\left(N + 1\right) \times D}$.\footnote{Here, we consider the single head case for simplicity. For more details, please refer to the original paper~\cite{vaswani2017transformer}.}
The outputs of the attention layer are fed to the following MLPs, which are composed of two linear layers with a non-linear activation between them and output tokens with their shape preserved.

Note that, as all transformer layers constituting the transformer encoder share the identical architecture, the final outputs from the ViT have the same shape as the input tokens, i.e., $\mathbb{R}^{\left(N + 1\right) \times D}$. For image classification task, only the {\fontfamily{qcr}\selectfont [CLS]} $\in \mathbb{R}^{D}$ is taken from the outputs and fed to a linear classifier. In our work, however, we consider the patch tokens $\mathbf{F}_{\text{Transformer}} \in \mathbb{R}^{N \times D}$ which are reshaped to $\mathbb{R}^{h \times w \times D}$ where $h$ and $w$ equal $\frac{H}{P} \times \frac{W}{P}$. It is worth noting that, the patch size $P$ plays the role of total stride as in the CNNs.\\

\begin{figure*}[t]
    \centering
    \includegraphics[width=.96\textwidth]{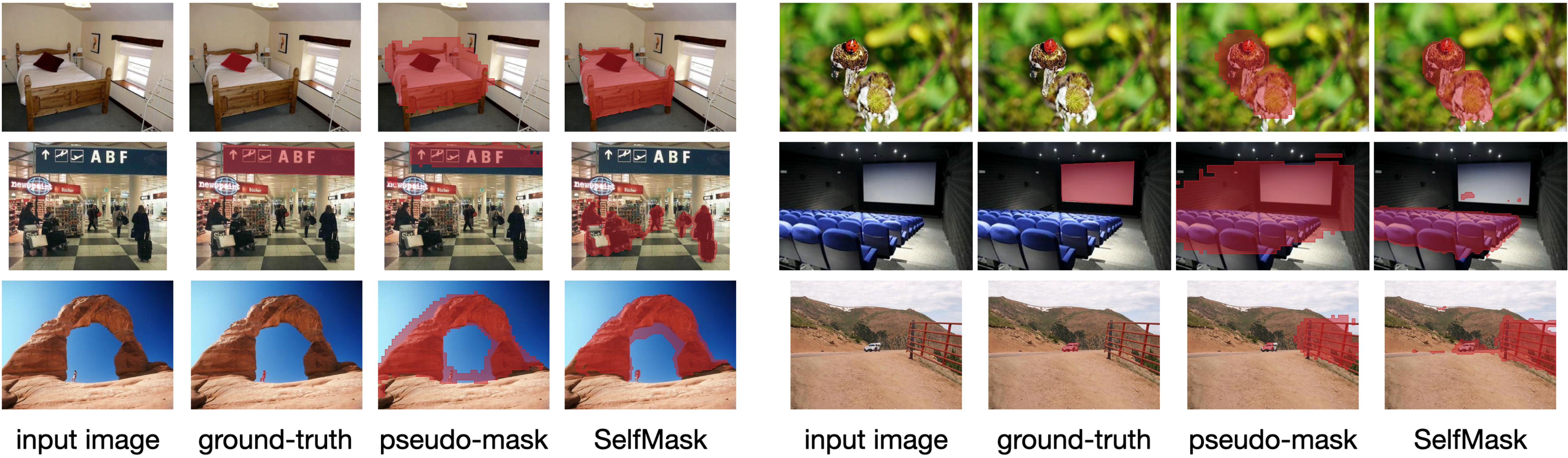}
    \caption{Sample visualisations for typical prediction failures from our model on the DUT-OMRON~\cite{yang2013saliency} and DUTS-TE~\cite{wang2017learning} benchmarks. From left to right, input image, ground-truth mask, a pseudo-mask, and a predicted mask by our model are shown.
    The respective salient regions are highlighted in red.
    Best viewed in colour. Please zoom in for details.}
    \mbox{}\\
    \label{supp:fig:failure-cases}
\end{figure*}

\section{Descriptions of evaluation metrics}\label{supp:metrics}
In the following, we describe the metrics used for evaluation:
\begin{itemize}
    \item $F_\beta$~\cite{Perazzi2012cvpr} is the
    harmonic mean of precision and recall
    between a ground-truth $G \in \{0, 1\}^{H\times W}$
    and a binarised mask $M \in \{0, 1\}^{H\times W}$:
    \begin{equation}
        F_\beta = \frac{(1+\beta^2)\text{Precision}\times \text{Recall}}{\beta^2 \text{Precision} + \text{Recall}},
    \end{equation}
    where $\beta^2$ denotes a weight of
    precision.\footnote{Precision$=\frac{tp}{tp + fp}$ and
    Recall $=\frac{tp}{tp + fn}$ where $tp$, $fp$,
    and $fn$ represent true-positive, false-positive,
    and false-negative, respectively.}
    Following previous work~\cite{wang2019sodsurvey, melas2021finding, zhao2019ICCV, Liu2019CVPR},
    we set $\beta^2$ to 0.3, putting more weight on precision. We use $F_\beta$ to compute the maximal-$F_\beta$, described next.
    \item maximal-$F_\beta$ (max$F_\beta$) is a maximum score
    of $F_\beta$ among multiple masks binarised with different
    thresholds.
    Specifically, given a non-binarised mask prediction with its
    value between [0, 255],
    it computes $F_\beta$ from 255 binarised masks, each of which
    is thresholded by an integer among $\{0, ..., 254\}$ and takes the
    maximum $F_\beta$ value for the result. 

    \item Intersection-over-union (IoU) is the size of overlapped foreground regions between a ground-truth $G$ and a binarised mask prediction $M$ divided by the total size of foreground regions from $G$ and $M$.
    \item Accuracy (Acc) is a metric that measures pixel-wise accuracy based on a ground-truth mask $G$ and a binarised mask prediction $M$:
    \begin{equation}
        \text{Acc} = \frac{1}{H \times W}\sum_{i=1}^{H}\sum_{j=1}^{W}\delta_{G_{ij}, M_{ij}}
    \end{equation}
    where $\delta$ denotes the Kroneker-delta.
\end{itemize}

\section{Comparison between \kmeans and spectral clustering}
\label{supp:kmeans-vs-spectral}

In Sec.~4.3 of the main paper, we show the performance of
\kmeans and spectral clustering applied to different architectures
(i.e., ResNet50 and ViT-S/\{8, 16\}) and features
(i.e., fully- and self-superivsed features) averaged over
$k$=$\{2, 3, 4\}$ on the three saliency datasets.
Here, we show the full results for each $k$
in Tab.~\ref{supp:tab:kmeans-vs-spectral}. %
For the description, please refer to Sec.~4.3 of the main paper.

\section{Visualisation of failure cases}
\label{supp:visualsations}
In Fig.~\ref{supp:fig:failure-cases},
we visualise some failure predictions from our model on the DUT-OMRON~\cite{yang2013saliency} and DUTS-TE~\cite{wang2017learning} datasets. 

We notice there are two typical failure cases.
First, when a salient object is of small scale, 
the model tends to undersegment it and prefers the large salient object. For instance, 
as shown by the top left example in Fig~\ref{supp:fig:failure-cases}, 
the whole bed is segmented, rather than the pillow;
Second, when there are more than one salient region in the image,
our model may only segment one of them. For example, as shown by the middle right example in Fig~\ref{supp:fig:failure-cases}, 
both screen and seats can be thought of as a salient region while the model only highlights only the latter.
We conjecture that these cases are caused by a bias of the dataset (\ie, DUTS-TR~\cite{wang2017learning}) on which the model is trained. That is, the training images likely to contain large salient regions composed of either an object or objects sharing a semantic meaning, thus discouraging the model from predicting a small salient region or more than one object with different semantics even if all the objects can be regarded salient.

\end{appendices}

%% file: tables/k-means-vs-spectral-full-results.tex
\setlength\dashlinedash{0.5pt}
\setlength\dashlinegap{1.5pt}
\setlength\arrayrulewidth{0.3pt}
\setlength{\tabcolsep}{6pt}  %

\begin{table*}[t]
\small
  \centering
        \begin{adjustbox}{center}
        \begin{tabular}{ccc
        ccc:c|ccc:c|ccc:c}
        \hline
        \multirow{2}{*}{Model}
        &\multirow{2}{*}{Arch.}
        &\multirow{2}{*}{Cluster.}
        &\multicolumn{4}{c}{DUT-OMRON~\cite{yang2013saliency}}
        &\multicolumn{4}{c}{DUTS-TE~\cite{wang2017learning}}
        &\multicolumn{4}{c}{ECSSD~\cite{shi2015hierarchical}}\\
        \cline{4-15}
        &&
        &$k$=2&$k$=3&$k$=4&avg.
        &$k$=2&$k$=3&$k$=4&avg.
        &$k$=2&$k$=3&$k$=4&avg.\\
        \hline
        \multicolumn{15}{c}{Fully-supervised features}\\
        \hline
        ResNet~\cite{He2016} & ResNet50 & \kmeans
        &.311&.346&.355&\textbf{.337}
        &.345&.358&.360&\textbf{.354}
        &.461&.445&.425&\textbf{.444}\\
        ResNet~\cite{He2016} & ResNet50 & spectral
        &.258&.326&.346&.310
        &.297&.341&.343&.327
        &.424&.454&.432&.437\\
        \hdashline
        ViT~\cite{dosovitskiy2021vit} & ViT-S/16 & \kmeans
        &.335&.406&.440&\textbf{.394}
        &.349&.423&.460&\textbf{.411}
        &.505&.560&.562&.542\\
        ViT~\cite{dosovitskiy2021vit} & ViT-S/16 & spectral
        &.268&.392&.481&.380
        &.260&.428&.511&.400
        &.402&.613&.637&\textbf{.551}\\
        \hline
        \multicolumn{15}{c}{Self-supervised features}\\
        \hline
        MoCov2~\cite{he2020momentum} & ResNet50 & \kmeans
        &.334&.387&.403&.375
        &.401&.423&.422&.415
        &.507&.511&.481&.500\\
        MoCov2~\cite{he2020momentum} & ResNet50 & spectral
        &.311&.399&.453&\textbf{.387}
        &.403&.464&.496&\textbf{.454}
        &.602&.642&.638&\textbf{.627}\\
        \hdashline
        SwAV~\cite{Caron20} & ResNet50 & \kmeans
        &.356&.412&.429&.399
        &.415&.456&.462&.444
        &.548&.552&.526&.542\\
        SwAV~\cite{Caron20} & ResNet50 & spectral
        &.346&.407&.450&\textbf{.401}
        &.412&.473&.488&\textbf{.458}
        &.594&.606&.569&\textbf{.590}\\
        \hdashline
        DINO~\cite{Caron_2021_ICCV} & ViT-S/8 & \kmeans
        &.299&.381&.427&.369
        &.299&.385&.447&.377
        &.497&.566&.591&.551\\
        DINO~\cite{Caron_2021_ICCV} & ViT-S/8 & spectral
        &.315&.417&.463&\textbf{.398}
        &.311&.435&.486&\textbf{.411}
        &.527&.616&.618&\textbf{.587}\\
        \hdashline
        DINO~\cite{Caron_2021_ICCV} & ViT-S/16 & \kmeans
        &.314&.391&.426&.377
        &.325&.407&.444&.392
        &.507&.557&.560&.541\\
        DINO~\cite{Caron_2021_ICCV} & ViT-S/16 & spectral
        &.310&.413&.459&\textbf{.394}
        &.324&.445&.483&\textbf{.417}
        &.528&.609.&.596&\textbf{.577}\\
        \hline
        \end{tabular}
\end{adjustbox}
\caption{Comparison between \kmeans algorithm and spectral clustering with three different cluster sizes $k$ on the three benchmarks. IoU between a ground-truth mask and the closest prediction among $k$ predicted masks is considered. On the fourth column of each benchmark, we report the average of the results from the different $k$. The higher average scores of \kmeans and spectral clusterings within the same model are in bold.
}
\label{supp:tab:kmeans-vs-spectral}
\end{table*}